# Inverting Variational Autoencoders for Improved Generative Accuracy

**Ian Gemp**  IMGEMP@CS.UMASS.EDU
**Mario Parente**  MPARENTE@ECS.UMASS.EDU
**Ishan Durugkar**  ISHAND@CS.UTEXAS.EDU
**Darby Dyar**  MDYAR@MTHOLYOKE.EDU
**Sridhar Mahadevan**  MAHADEVA@CS.UMASS.EDU

## Abstract

Recent advances in semi-supervised learning with deep generative models have shown promise in generalizing from small labeled datasets $(\mathbf{x}, \mathbf{y})$ to large unlabeled ones $(\mathbf{x})$. In the case where the codomain has known structure, a large un*featured* dataset $(\mathbf{y})$ is potentially available. We develop a parameter-efficient, deep semi-supervised generative model for the purpose of exploiting this untapped data source. Empirical results show improved performance in disentangling latent variable semantics as well as improved discriminative prediction on Martian spectroscopic and handwritten digit domains.

## 1. Introduction

Semi-supervised learning aims to improve learning accuracy when a large source of unlabeled data $(\mathbf{x}_u)$ is available in addition to a small labeled dataset $(\mathbf{x}_l, \mathbf{y}_l)$. Under this setting, inductive learning specifically judges accuracy of the learned mapping $f : \mathbf{x} \to \mathbf{y}$ for all $\mathbf{x}$, while transductive learning focuses on accuracy of this mapping for only $\mathbf{x}_u$. Semi-supervised techniques have significantly improved inductive and transductive learning accuracy for applications ranging from website classification (Blum & Mitchell, 1998), to natural language processing (Turian et al., 2010), to image segmentation and search (Fergus et al., 2009; Papandreou et al., 2015).

Separately, deep probabilistic models leveraging advances in variational inference have made gains in modeling text (Miao et al., 2016), images (Gulrajani et al., 2016), and speech (Chung et al., 2015). The architecture common to these models is a deep generative model deemed the *variational autoencoder* (Kingma & Welling, 2013). Kingma et al. (2014) developed an extension of this architecture to semi-supervised tasks with excellent semi-supervised classification performance on MNIST (LeCun et al., 1998).

One advantage of using generative models for semi-supervised learning is the availability of a mechanism for generating unobserved data conditioned on the labels. In addition to generating novel observations, this mechanism can be used for exploring the data manifold and typically reveals interesting structure suggesting semantics are disentangled during training.

While semi-supervised learning typically focuses on inductive and transductive learning, in this work, we also stress the importance of generative accuracy ($f^{-1} \sim g : \mathbf{y} \to \mathbf{x}$), which is of growing interest in generative modeling (e.g., GANs (Goodfellow et al., 2014)).

It is the goal of generative accuracy that leads us to explore the possibility of identifying and exploiting a large source of un*featured* data $(\mathbf{y}_u)$: $\mathbf{y}$ without corresponding $\mathbf{x}$. In general, if 1) we know the support of $\mathbf{y}$, 2) we have a strong prior belief on $p(\mathbf{y})$, and 3) label instances can be easily synthesized, then $\mathbf{y}_u$ constitutes a potential data source to be tapped. In this work, we consider two different supports for $\mathbf{y}$: a discrete finite set and a simplex. In each domain that we explore, we are able to exploit our prior knowledge of $\mathbf{y}$ to improve generative accuracy and in some cases discriminative accuracy as well.

Our approach is essentially to "invert" the deep generative model used for standard semi-supervised learning and train it in reverse, thereby considering the flipped semi-supervised task of learning $g : \mathbf{y} \to \mathbf{x}$ given $(\mathbf{x}_l, \mathbf{y}_l)$ and $\mathbf{y}_u$. A key feature of our approach is that $f$ and $g$ are learned jointly with tied parameters to better regularize the model.

For the sake of clarity, we first describe a problem for which an untapped source of labels indeed exists. We then describe our model within this context. Finally, we continue with a twist on a more familiar dataset, MNIST, exemplifying the generality of our approach.

## 2. Spectroscopic Analysis

The *Curiosity* rover has been exploring the Martian surface for several years now, during which time its laser-induced breakdown spectroscopy (LIBS) instrument

ChemCam (Wiens et al., 2013; Clegg et al., 2016) has been transmitting daily spectroscopic data from dust, soil, and rocks. Differences between Mars and Earth surface conditions such as atmospheric pressure and plasma temperature significantly affect the spectroscopic data, such that spectra received from Mars cannot necessarily be matched up with spectra taken from samples with known elemental composition on Earth. While the vast majority of the spectra taken from the Martian surface essentially arrive at Earth unlabeled (with unknown compositions), a small labeled 10-sample dataset of standards is available in a calibration target assembly attached to the rover. Spectra are regularly acquired from the rock samples in the target assembly to produce a set of ground truth data. Although the standards were chosen by NASA scientists to be representative of geochemical samples encountered on Mars, they represent only a small subset of what the rover is expected to investigate. Therefore, a major goal of this data analysis effort is to predict the compositions of samples that exist outside the set of standards. To expand the range of predictive accuracy, we need to take advantage of data from sources beyond the ground truth palette. In this study, we use spectroscopic data collected in the lab under Martian atmospheric conditions to simulate the rover environment. Specifically, we use 88 spectra from the same standards found on Curiosity's calibration target and 500 spectra from unlabeled rocks and soils typically found on the Martian surface.

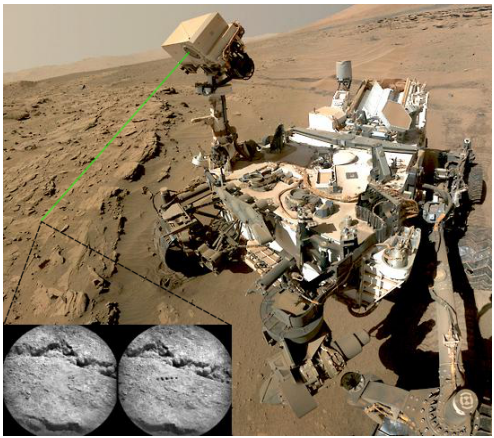

Figure 1. The *Curiosity* rover on Mars with a simulated ChemCam laser pulse. The photos on the left are of a Martian rock surface before and after laser ablation (LIBS shot). The rock was lased 50 times in each of the five locations. Photos courtesy of NASA.

The *Compact Reconnaissance Imaging Spectrometer* (CRISM) aboard the *Mars Reconnaissance Orbiter* captures hyperspectral images of the Martian surface from space, providing an essential orbital counterpart to the ChemCam's fine grained data capture on the surface. Like LIBS, CRISM also suffers from observation noise and unpredictable environmental conditions. Unlike LIBS, CRISM is moving relative to its target during measurement and covers a much larger area ($\sim$18m /pixel), so that different pixels are acquired at varying observational geometries (i.e. incidence angles) and the recorded spectra can be considered as complex non-linear (*intimate*) mixtures of the spectra of the pure minerals present in the scene, referred to as *endmembers*.

To simulate CRISM conditions, we consider the following laboratory experiment. We construct a dataset with samples from particulate mixtures of three endmembers: a forsteritic olivine, a diopside, and a bytownite feldspar. The abundance[1] simplex for their relative proportions was sampled with a regular grid of 66 points (mineral ratios) for which we created intimately mixed samples by weighing out an appropriate quantity of each of the endmembers. We then imaged all the 66 mixture samples using a Micro-Hyperspec® SWIR M-Series imaging sensor to retrieve sample reflectance. Each sample was imaged at three fixed geometries (referred to as "Position 1", "Position 2" and "Angle 2") to simulate the dynamic geometry of CRISM relative to its target. Figure 2 illustrates the three configurations. In Position 1, the camera is perpendicular to the sample and the illumination source is at $\approx$30 degrees. In Position 2, the illumination is shifted away from the sample. The configuration denoted Angle 2 is obtained by tilting the sample compartment with respect to the sample-camera line and with illumination source in Position 2.

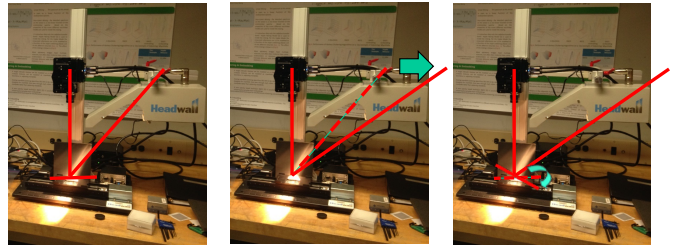

Figure 2. Acquisition geometries. Left (Position 1), center (Position 2), right (Angle 2)

A primary concern of hyperspectral imaging is *spectral unmixing* in which each spectrum (high dimensional pixel) is identified as a mixture of constituent endmembers (e.g., olivine + feldspar) with corresponding abundance. This task is challenging because endmember spectra (pure minerals) are typically not present in the environment because rocks are finely-mixed assortments of minerals. Given an accurate generative model of the mixing process, spectral unmixing is accomplished by generating spectra condi-tioned on an abundance corresponding to a pure endmember sample (corner of the simplex).

---

[1] We will use abundance and composition interchangeably.

The ability to generate spectra (conditioned on compositions/abundances) also allows scientists to investigate the spectra of hypothetical samples not yet encountered on the Martian surface. Therefore, both aspects of the generative model, the discriminative prediction of spectral abundance ($f$) and the generation of unseen spectra ($g$), are needed for spectroscopic analysis.

### 2.1. Compositions & Abundances Untapped

In the case study described above, we know that any mixture of endmembers is legitimate. In other words, any $\mathbf{y}$ on the simplex can be considered a viable label. Furthermore, if the goal is unmixing, we are primarily concerned with the corners of the simplex because those are the data that correspond to pure endmembers. We ought to be able to exploit this knowledge to better learn the inductive, $f$, and generative, $g$, mappings.

We appeal to intuition in the context of a vanilla autoencoder (see Figure 3) and carry over insights developed in that setting to the variational model where distributions replace point estimates. Although variational autoencoders have been shown to behave differently from vanilla autoencoders (specifically with regards to representation learning), our results suggest the insights gleaned are valuable.

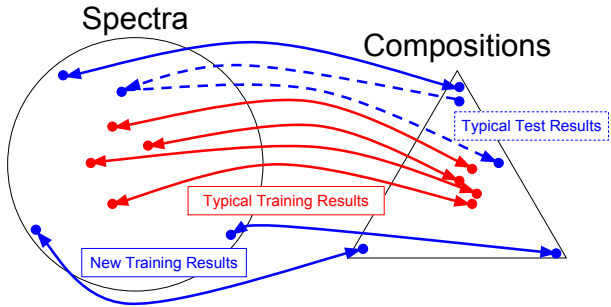

*Figure 3.* Typical autoencoder training roughly learns a one-to-one mapping only over the training set (solid red). When the model is introduced to new compositions (e.g., corners) and asked to decode them, its predictions often map back to compositions different from the original input (dashed, blue)! With our new approach, the model explicitly receives a signal to build a one-to-one mapping for compositions outside the training set (solid blue).

## 3. Deep Semi²-Supervised Generative Model

Motivated by the above problem, we consider the marriage of two probabilistic models to describe the data. The first is a probabilistic model (M2 in (Kingma et al., 2014)) that describes the spectra as being generated by a composition vector $\mathbf{y}$ in addition to a latent, nuisance vector $\mathbf{z}$. The joint distribution is assumed to factorize as $p(\mathbf{x}, \mathbf{y}, \mathbf{z}) = p(\mathbf{y})p(\mathbf{z})p(\mathbf{x}|\mathbf{y}, \mathbf{z})$, so the data are explained by the *generative process*:

$$p(\mathbf{y}) = Dir(\mathbf{1}) \quad (1)$$
$$p(\mathbf{z}) = U(-\mathbf{1.5}, \mathbf{1.5}) \quad (2)$$
$$p_\theta(\mathbf{x}|\mathbf{y}, \mathbf{z}) = f(\mathbf{x}; \mathbf{y}, \mathbf{z}, \theta) \quad (3)$$

Here, $p(\mathbf{y})$ and $p(\mathbf{z})$ are prior distributions and $f(\mathbf{x}; \mathbf{y}, \mathbf{z}, \theta)$ is a distribution whose parameters are non-linear functions of $\mathbf{y}$ and $\mathbf{z}$ (e.g., diagonal Gaussian $\mathcal{N}(\mu_\theta(\mathbf{y}, \mathbf{z}); \Sigma_\theta(\mathbf{y}, \mathbf{z}))$). We choose a uniform prior over the simplex for compositions, $Dir(\mathbf{1})$, and deep neural networks with weights $\theta$ for $\mu_\theta(\mathbf{y}, \mathbf{z})$ and $\Sigma_\theta(\mathbf{y}, \mathbf{z})$.

The second is a probabilistic model that describes the *reverse process*: nuisances and compositions are generated by spectra,

$$q(\mathbf{x}) = U(-\gamma, \gamma) \text{ e.g., } \gamma \gg 0 \quad (4)$$
$$q_\phi(\mathbf{y}|\mathbf{x}) = g(\mathbf{y}; \mathbf{x}, \phi) \quad (5)$$
$$q_\phi(\mathbf{z}|\mathbf{x}, \mathbf{y}) = h(\mathbf{z}; \mathbf{x}, \mathbf{y}, \phi) \quad (6)$$

where $q(\mathbf{x})$ is an uninformative, uniform prior, $q_\phi(\mathbf{z}|\mathbf{x}, \mathbf{y})$ is a diagonal logistic-normal distribution parameterized a deep neural network as before, and $\gamma \gg 0$. While $q_\phi(\mathbf{z}|\mathbf{x}, \mathbf{y})$ is unnecessary for generating $\mathbf{y}$, our reverse model formulation is actually a specific instance of an *auxiliary* generative model (Maaløe et al., 2016), which was shown to make the variational distribution more expressive. This term also serves an additional role described next.

Computing the exact posterior of the latent variables $\mathbf{y}$ and $\mathbf{z}$ (i.e., Bayesian inference) in the first model, and likewise $\mathbf{x}$ in the second model, is intractable due to non-conjugate ($Dir$ & $U$), non-linear (deep nets) dependencies. Instead, we approximate the posterior distribution with a separate non-linear function called a *recognition model* for inferring or "recognizing" the latent variables. One of our novel contributions is to reuse $q_\phi(\mathbf{y}|\mathbf{x})q_\phi(\mathbf{z}|\mathbf{x}, \mathbf{y})$ as the recognition model for the forward generative model and $p(\mathbf{z})p_\theta(\mathbf{x}|\mathbf{y}, \mathbf{z})$ for the reverse model.

To learn the parameters, $\theta$ and $\phi$, we optimize variational lower bounds on the marginal likelihoods of our data samples. Lower bounds for the forward model are given by

$$\log p_\theta(\mathbf{x}, \mathbf{y}) \geq \mathcal{L}_{fxy} = \mathbb{E}_{q_\phi(\mathbf{z}|\mathbf{x}, \mathbf{y})} \Big[ \log p_\theta(\mathbf{x}|\mathbf{y}, \mathbf{z})$$
$$- \log q_\phi(\mathbf{z}|\mathbf{x}, \mathbf{y}) + \log p(\mathbf{y}) + \log p(\mathbf{z}) \Big] \quad (7)$$

$$\log p_\theta(\mathbf{x}) \geq \mathcal{L}_{fx} = \mathbb{E}_{q_\phi(\mathbf{y}, \mathbf{z}|\mathbf{x})} \Big[ \log p_\theta(\mathbf{x}|\mathbf{y}, \mathbf{z})$$
$$- \log q_\phi(\mathbf{y}, \mathbf{z}|\mathbf{x}) + \log p(\mathbf{y}) + \log p(\mathbf{z}) \Big] \quad (8)$$

respectively for labeled and unlabeled samples.

Lower bounds for the reverse model are given by

$$\log q_\phi(\mathbf{x}, \mathbf{y}) \geq \mathcal{L}_{rxy} = \mathbb{E}_{p(\mathbf{z})}\Big[\log q_\phi(\mathbf{y}|\mathbf{x})$$
$$+ \log q_\phi(\mathbf{z}|\mathbf{x}, \mathbf{y}) - \log p(\mathbf{z}) + \log q(\mathbf{x})\Big] \quad (9)$$

$$\log q_\phi(\mathbf{y}) \geq \mathcal{L}_{ry} = \mathbb{E}_{p(\mathbf{z})p_\theta(\mathbf{x}|\mathbf{y},\mathbf{z})}\Big[\log q_\phi(\mathbf{y}|\mathbf{x})$$
$$+ \log q_\phi(\mathbf{z}|\mathbf{x}, \mathbf{y}) - \log p(\mathbf{z}) - \log p_\theta(\mathbf{x}|\mathbf{y}, \mathbf{z})$$
$$+ \log q(\mathbf{x})\Big] \quad (10)$$

respectively for labeled and un*featured* samples.

The marginal likelihood for the entire dataset is then

$$\mathcal{J}_f = \sum_{(\mathbf{x},\mathbf{y})\sim\tilde{p}_l} (\mathcal{L}_{fxy}) + \sum_{\mathbf{x}\sim\tilde{p}_{u_x}} (\mathcal{L}_{fx}) \quad (11)$$

$$\mathcal{J}_r = \sum_{(\mathbf{x},\mathbf{y})\sim\tilde{p}_l} (\mathcal{L}_{rxy}) + \sum_{\mathbf{y}\sim\tilde{p}_{u_y}} (\mathcal{L}_{ry}) \quad (12)$$

for the forward and reverse models, respectively.

The predictive distribution for compositions appears in 8, 9, and 10, but not 7. Likewise, the generative distribution for spectra appears in 7, 8, and 10, but not 9. As in (Kingma et al., 2014), we introduce an additional discriminative objective to each model that can be learned from the labeled data:

$$\mathcal{J}_f^d = \mathbb{E}_{(\mathbf{x},\mathbf{y})\sim\tilde{p}_l} L(\bar{\mathbf{y}}, \mathbf{y}) \quad (13)$$
$$\mathcal{J}_r^d = \mathbb{E}_{(\mathbf{x},\mathbf{y})\sim\tilde{p}_l} L(\bar{\mathbf{x}}, \mathbf{x}) \quad (14)$$

where $\bar{\mathbf{y}}$ can, for example, either be a sample from $q_\phi(\mathbf{y}|\mathbf{x})$ or the mean of the distribution and $L$ can, for example, be KL($\mathbf{y} \parallel \bar{\mathbf{y}}$); $\mathbf{x}$ is treated similarly using $p_\theta(\mathbf{x}|\mathbf{y}, \mathbf{z})$ and $L = \parallel \bar{\mathbf{x}} - \mathbf{x} \parallel^2$.

This still leaves the question of how to combine both models. We could introduce an additional latent variable that interpolates between both models, capturing our uncertainty in the nature of the generative process: are spectra a result of composition or vice versa? Alternatively, we could specify a joint prior over the weights of both models as in (Shu et al., 2016). Instead, we take the view from multi-objective optimization and opt to weight each objective with a coefficient set by cross-validation:

$$\mathcal{J} = \alpha_f \mathcal{J}_f - \alpha_f^d \mathcal{J}_f^d + \alpha_r \mathcal{J}_r - \alpha_r^d \mathcal{J}_r^d. \quad (15)$$

Computing the probability densities of diagonal Gaussians and logistic-normals is trivial; sampling from diagonal Gaussians and logistic-normals is simple with a reparameterization trick:

$$\mathbf{x} \sim \mu_\theta(\mathbf{y}, \mathbf{z}) + \Sigma_\theta^{1/2}(\mathbf{y}, \mathbf{z}) \cdot \epsilon \quad (16)$$
$$\mathbf{y} \sim \texttt{softmax}(\mu_\phi(\mathbf{y}, \mathbf{z}) + \Sigma_\phi^{1/2}(\mathbf{y}, \mathbf{z}) \cdot \epsilon) \quad (17)$$

where $\epsilon$ is sampled from a standard multivariate normal distribution. To generate an endmember, we simply draw a sample from $p_\theta(\mathbf{x}|\mathbf{y}, \mathbf{z})$ where $\mathbf{y}$ is the corner of the simplex representing 100% of the corresponding mineral. Likewise, to infer the nuisance variable for a given spectrum, we draw a sample from $q_\phi(\mathbf{z}|\mathbf{x}, \mathbf{y})$, where $\mathbf{y}$ can be drawn from $q_\phi(\mathbf{y}|\mathbf{x})$ if it is not available.

We learn the parameters $\theta$ and $\phi$ by maximizing (15) using Monte Carlo samples for the latent variables —a technique known as stochastic gradient variational Bayes (Kingma & Welling, 2013) or stochastic backpropagation (Rezende et al., 2014). Pseudocode is given in Algorithm 1 where $\Gamma(g_\theta, g_\phi)$ returns a parameter update increment (e.g., $\Gamma = \text{SGD} \to -(g_\theta, g_\phi)$).

---

**Algorithm 1** Learning the Model

**while** training() **do**
  $\mathcal{D} \leftarrow$ getRandomMiniBatch()
  $\mathcal{J} = 0$
  **for all** $\{\mathbf{x}_i, \mathbf{y}_i\} \in \mathcal{D}_{\{x,y\}}$ **do**
    $\mathbf{z}_i \sim q_\phi(\mathbf{z}|\mathbf{x}_i, \mathbf{y}_i),\ \bar{\mathbf{y}}_i = \texttt{mean}(q_\phi(\mathbf{y}|\mathbf{x}_i))$
    $\mathcal{J}\mathrel{+}= \alpha_f(7) - \alpha_f^d L(\bar{\mathbf{y}}_i, \mathbf{y}_i)$
    $\mathbf{z}_i \sim p(\mathbf{z}),\ \bar{\mathbf{x}}_i = \texttt{mean}(p_\theta(\mathbf{x}|\mathbf{y}_i, \mathbf{z}_i))$
    $\mathcal{J}\mathrel{+}= \alpha_r(9) - \alpha_r^d L(\bar{\mathbf{x}}_i, \mathbf{x}_i)$
  **end for**
  **for all** $\mathbf{x}_i \in \mathcal{D}_x$ **do**
    $\mathbf{y}_i \sim q_\phi(\mathbf{y}|\mathbf{x}_i),\ \mathbf{z}_i \sim q_\phi(\mathbf{z}|\mathbf{x}_i, \mathbf{y}_i)$
    $\mathcal{J}\mathrel{+}= \alpha^f(8)$
  **end for**
  **for all** $\mathbf{y}_i \in \mathcal{D}_y$ **do**
    $\mathbf{z}_i \sim p(\mathbf{z}),\ \mathbf{x}_i \sim p_\theta(\mathbf{x}_i|\mathbf{y}_i, \mathbf{z}_i)$
    $\mathcal{J}\mathrel{+}= \alpha_r(10)$
  **end for**
  $(g_\theta, g_\phi) \leftarrow (-\frac{\partial \mathcal{J}^\alpha}{\partial \theta}, -\frac{\partial \mathcal{J}^\alpha}{\partial \phi})$
  $(\theta, \phi) \leftarrow (\theta, \phi) + \Gamma(g_\theta, g_\phi)$
**end while**

---

### 3.1. Generalizing the Model

We can generalize the formulation above to new problem domains by varying the priors, discriminative losses, normalizing flows, and/or the reparameterized distributions. For example, later on, in our MNIST experiment, we swap out the diagonal Gaussian ($\mu + \Sigma^{1/2}\epsilon$) for the *gumbel-softmax / concrete* distribution (Jang et al., 2016; Maddison et al., 2016) in order to model an approximately categorical distribution.

In the model above, we treat $\mathbf{z}$ as a latent variable in the reverse process. In experiments, we instead treat $\mathbf{z}$ as observed and include it as part of our *unfeatured* dataset, $(\mathbf{y}_u, \mathbf{z}_u)$, which reinforces our prior on $\mathbf{z}$. Another possible reformulation is to introduce an additional network, $q_\phi(\mathbf{z}|\mathbf{y}) \approx \int_\mathbf{x} q_\phi(\mathbf{z}|\mathbf{x}, \mathbf{y})$, to replace $p(\mathbf{z})$ as part of inference in the reverse process. The integral approximation could be encouraged by minimizing KL-divergence with Monte Carlo samples. The proposed model with $\mathbf{z}_u$ observed is implemented using the Theano (Theano Development Team, 2016), Lasagne, Parmesan, and Scikit-learn (Pedregosa et al., 2011) packages[2].

## 4. CRISM Experiment

We experiment with the proposed model in the lab setting designed to mimic conditions experienced by CRISM in Mars orbit. As mentioned earlier, we imaged the mineral mixtures with varying configurations to approximate the dynamic geometry of the Mars satellite relative to the Martian surface. The values of incidence and emission angle for the three configurations are reported in Table 1.

Table 1. Acquisition angles for the three imaging configurations.

|  | Incidence angle $\theta_i$ | Emission angle $\theta_e$ |
| --- | --- | --- |
| Position 1 | 33.9 | 0 |
| Position 2 | 42.7 | 0 |
| Angle 2 | 12.7 | 30 |

Overall, we extract 5354 spectra for all three configurations. Each spectrum is represented by a vector of 165 reflectance values between 0 and 1. Figure 4 represents the nominal abundance grid. We select 500 spectra with the corresponding abundance $(\mathbf{x}_l, \mathbf{y}_l)$ and 992 unlabeled spectra $\mathbf{x}_u$ from the set of spectra with nominal abundances represented by the triangles in Figure 4. This means a vast majority of the simplex (including the endmembers) is not seen by the model. We select 501 abundance vectors $\mathbf{y}_u$ corresponding to the corners of the abundance simplex (circles in Figure 4) to train the reverse model. We set $\mathbf{z}$ to be 1-dimensional.

### 4.1. Endmember Extraction

We find evidence of the model's successful performance in the visual inspection of the retrieved endmember spectra, depicted with thin, solid lines in Figure 5, together with the spectra of the true endmembers (thick, transparent). In fact, only the retrieved bytownite spectrum (first from the top) presents some deviation in spectral channels 1-60, with respect to the corresponding true spectrum. This likely occurs because that mineral has only a small signal owing to very low concentrations of iron, which is the element giving rise to the spectral features, especially at the lower wavelengths where the mismatch is greatest. Moreover, our proposed model reduces endmember extraction error over

---

[2]Code available @ https://github.com/all-umass/untapped

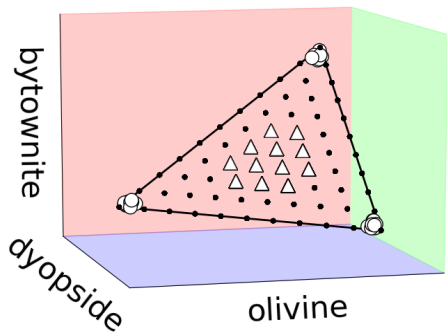

Figure 4. Abundance grid: A randomly selected subset of the triangles supplies data for $(\mathbf{x}_l, \mathbf{y}_l)$ while the entire set of triangles serves as $\mathbf{x}_u$. $\mathbf{y}_u$ is randomly drawn from samples near the corners of the simplex (circles).

M2 (Kingma et al., 2014) by $\approx 13\%$.

### 4.2. Abundance Prediction

Partial least squares (PLS) is well known in the spectroscopic community and functions as the de facto multivariate standard for predicting compositions (Wold et al., 2001). PLS is especially effective when features in the training set are collinear. Different mineral or rock mixtures are often identified by spectral signatures at specific channel bands, so this condition is often approximately met. We use PLS here (see Table 2) as a strong baseline to compare against. Although PLS outperforms both the VAE models, it is important to note that introducing *unfeatured* $\mathbf{y}$ into the VAE framework improves endmember extraction without degrading composition prediction (relative to M2).

Table 2. Validation Error: KL-divergence from predicted distribution over endmembers to ground truth.

|  | PLS | M2 | Untapped |
| --- | --- | --- | --- |
| KL | 0.036 | 0.061 | 0.058 |

### 4.3. Latent Semantics

We visualize the nuisance value for each spectrum in the training set in Figure 6a. The values of $\mathbf{z}$ are sorted according to the configuration that the corresponding spectra were acquired at. We can see that the nuisance median value fluctuates around three sharply different levels (grey horizontal lines) corresponding to different configurations. The actual level values are inversely proportional to the variations in incidence angle in Table 1 ( $\theta_{i,\text{Pos2}} - \theta_{i,\text{Pos1}} =$

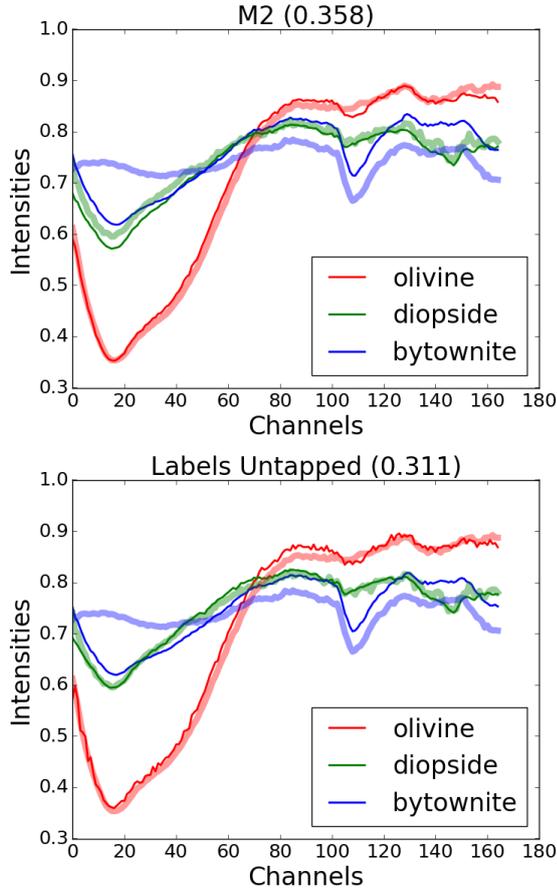

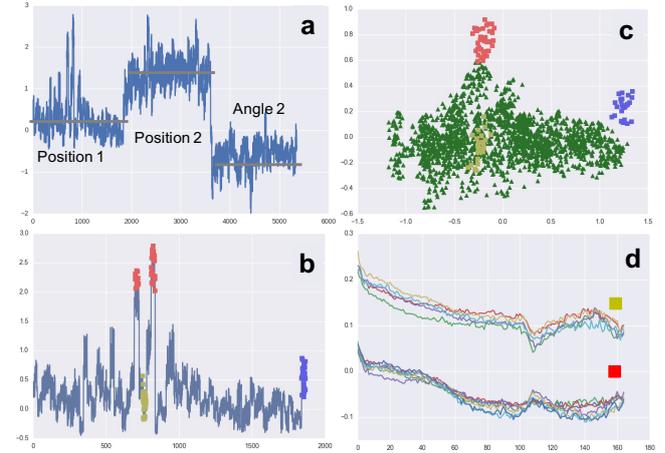

*Figure 5.* Mineral endmembers: Training our proposed model with $\mathbf{y}_u$ from the corners of the simplex reduces endmember error ($\| \bar{\mathbf{x}} - \mathbf{x} \|_2$) over M2 (Kingma et al., 2014) from 0.358 to 0.311. Qualitatively, our approach better matches diopside at lower wavelengths ($0 - 40$) and separates the signals at higher wavelengths ($100 - 120$).

$42.7 - 33.9 = 8.8 < \theta_{i,\text{Pos2}} - \theta_{i,\text{Angle2}} = 42.7 - 12.7 = 30$).

The small variations of $\mathbf{z}$ around each configuration level could be ascribed to variations in grain size and the randomness of the true abundance vectors. In fact, these mineral samples exhibit a distribution of grain sizes in the range 63-108 $\mu$m. Furthermore, the endmember abundance values for each mixed sample all vary slightly about the nominal values depicted in the abundance grid (Figure 4). All pixel spectra of samples with the same nominal abundances are, in a way, sampling from the distributions of grain sizes and of the true abundances. The small fluctuations in nuisance values are consistent with the small difference in reflectance for such pixels. The nuisance variable is also able to capture outliers. Figure 6b shows an enlargement of the nuisance signal in Figure 6a corresponding to spectra acquired in Position 1. Figure 6c shows the complete spec-tral dataset projected onto the first three principal components. The red (and blue) squares in Figure 6a highlight outlying values of the nuisance variable corresponding to the similarly colored clear spectral outliers in Figure 6c. As a sanity-check, we verify the correspondence for non-outlying points (in gold). For further confirmation, we plot in Figure 6d the spectra for the points in the red and gold clusters of Figure 6c. Gold spectra represent legitimate mixed spectra while the red spectra are pixels that erroneously capture the surface of the sample container.

*Figure 6.* Characterization of the nuisance. In **a**, the nuisance signal. In **b**, the nuisance signal in Position 1. Red squares: nuisance values corresponding to similarly colored outlier points in **c**. In gold, nuisance values corresponding to similarly-colored non-outlying points in **c**. Depicted in **d** are the plots of the spectra in red and gold from **c**.

## 5. LIBS Experiment

In this section, we examine the LIBS spectra received from the ChemCam[3] instrument onboard the Mars rover *Curiosity*. LIBS spectra reflectance is recorded at 6144 channels, then reduced to 5485 channels after removal of bad channels. LIBS spectra are dominated by emission lines from nine major elements, introducing an increase in complexity over the previous mineral model. The rest of the setup is the same. However, we no longer have fine-grained control or knowledge of instrumental factors, which in the case of LIBS, include plasma distribution and laser coupling.

### 5.1. Endmember Extraction

As before, we train our proposed model and use the generative (decoder) network to extract endmembers. Figure 7 reveals the spectral signal generated from setting $\mathbf{y}$ to

---
[3] http://pds-geosciences.wustl.edu/msl/msl-m-chemcam-libs-4_5-rdr-v1/

100% $SiO_2$. The emission lines correspond to elemental Si, though the geological convention is to express Si concentration in terms of its oxide component, $SiO_2$. It confirms a strong agreement between the elemental signal generated by the model and the known spectral emission features.

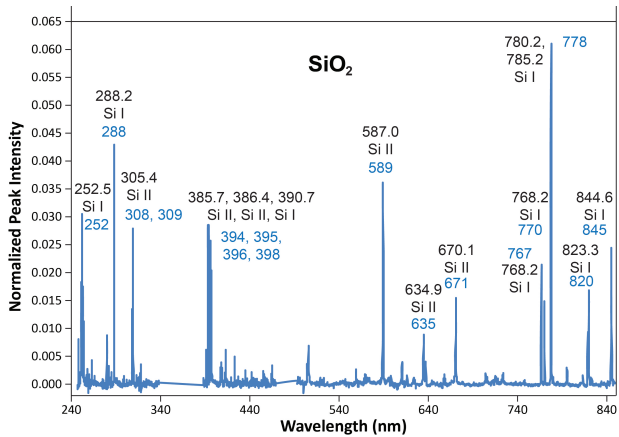

*Figure 7.* Spectral signal generated from our model conditioned on a composition of pure $SiO_2$. Annotations in blue denote the locations of peaks in the generated signal while black denotes the location of peaks known to correspond to emission lines of Si identifying $SiO_2$.

### 5.2. Composition Prediction

There are only ten individual standards on the calibration target secured to the Mars rover. Thus our only labeled dataset consists of repeated shots at each of the standards. Moreover, the standards themselves can only sustain so many shots until the laser ablates and redeposits the material so the labeled dataset may be changing over time. Measurements taken of two of the standards (Macusanite, KGA-Med) are poor, and we have only six samples total for two others (Graphite, Titanium), therefore, our dataset is reduced to six standards. Of the remaining six standards, Norite, Picrite, and Shergottite are synthetic mixtures of oxides made to resemble a Martian meteorite composition, and Nau2-(Lo,Med,Hi) is a rock powder mixed with varying amounts of sulfur. If we were to randomly shuffle this data and perform validation (k-fold or stratified CV), we would expect to find that our deep model can learn an accurate mapping from spectra to compositions by overfitting the data because repeated shots of the same standard are similar. In addition, we stated earlier the desire for our model to predict compositions of samples outside the standard dataset. To this end, we perform random shuffling at the compound level so that we are training on a subset of the compounds (e.g., 3/6) and testing on the remaining set (e.g., 3/6). This way, we can better test generalization error.

*Table 3.* Validation Error: Predicting compositions of spectra from rock types $\in$ {Nau2-Lo, Nau2-Med, Nau2-Hi} (88 samples) by training on {Norite, Picrite, Shergottite} (42 samples). Note that this task (leave-p-out) is more difficult than a stratified sampling approach, but important in order to simulate *Curiosity*'s encounter with alien rocks. Error is given by KL-divergence from predicted composition to ground truth.

|    | PLS   | M2    | Untapped |
|----|-------|-------|----------|
| KL | 0.922 | 0.618 | 0.525    |

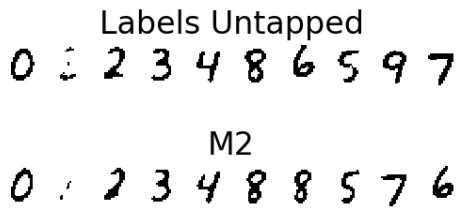

*Figure 8.* Digit Extraction (means): Our proposed model (top) and M2 (bottom) generate images (means, $\mu$) conditioned on one-hot vectors ($\mathbf{y}$). The models are provided a supervised signal to assign digits $0-4$ to the first 5 dimensions. No signal is given for assigning the remaining dimensions. Notice, (top) assigns a unique digit to each dimension, while the digit 9 does not appear in (bottom).

Table 3 reveals significant improvement in the VAE model predictions over PLS. Of the two VAEs, our proposed model shows lower generalization error.

## 6. MNIST Experiment

MNIST digit recognition provides a well-known benchmark in machine learning for supervised, semi-supervised, and unsupervised learning. Here, we construct a new MNIST task analogous to the spectroscopic tasks above. Specifically, we present a semi-supervised variant where labeled data is only available for a subset of the digits, $0-4$. The model is not only expected to attribute correct semantics to $\mathbf{y}$ for digits $0-4$, but also $5-9$, which makes this task particularly challenging. In other words, as before, the model should be able to accurately generate unseen observations conditioned on one-hots (e.g., $\mathbf{y} = [0,\ldots,1]$). Note that while the model is provided a signal to attribute the first five dimensions of $\mathbf{y}$ to $0-4$, the remaining dimensions are free to disentangle meaning as the model sees fit; the hope is that $5-9$ will be ascribed to some random permutation of the remaining dimensions and not distributed across the dimensions (e.g., $5 \neq [1/10,\ldots,1/10])$, folded into $\mathbf{z}$, or stored in the weights of the network. We use a 2d $\mathbf{z}$ as in (Kingma & Welling, 2013) to capture variation in the digits.

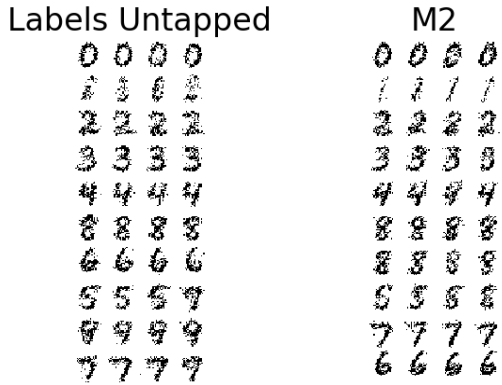

*Figure 9.* Digit Extraction (samples): Our proposed model (left) and M2 (right) generate images ($\mathbf{x} \sim p_\theta(\mathbf{x}|\mathbf{y}, \mathbf{z})$ conditioned on one-hot vectors ($\mathbf{y}$). The models are provided a supervised signal to assign digits $0 - 4$ to the first 5 dimensions. No signal is given for assigning the remaining dimensions. Notice, (left) assigns a unique digit to each dimension, while the digit 8 appears in rows $6 - 8$ of (right).

Figures 8 and 9 demonstrate the tendency of our proposed model to encourage $\mathbf{y}$ towards the desired representation. All digits are generated by conditioning on the 10 possible one-hots and the mean of $\mathbf{z}$ over the training set:

$$\mathbf{x}' \sim p_\theta(\mathbf{x}|\mathbf{y} = [1, \ldots, 0], \bar{\mathbf{z}} = \mathtt{mean}(q_\phi(\mathbf{z}|\mathbf{x_l}, \mathbf{y_l})). \quad (18)$$

In the M2 model, the digit 8's representation is distributed across the sixth, seventh, and partially eighth dimensions of $\mathbf{y}$. In fact, the digit 9 does not appear to be represented by any one-hot representation in the M2 model. In contrast, our proposed model has attributed a unique digit to each possible one-hot vector: $0, 1, 2, 3, 4|\mathbf{8, 6, 5, 9, 7}$.

### 6.1. Digit Prediction

It is not surprising, given our model's learned representation, that training with $\mathbf{y}_u$ results in improved generalization performance. Table 4 lists KL-divergence from the predicted distribution over digits to the true one-hot distribution. Training with *unfeatured* $\mathbf{y}$ reduces the average distance to ground truth by $30\%$.

*Table 4.* Validation Error: KL-divergence from predicted distribution over digits to ground truth.

|    | M2    | Untapped |
|----|-------|----------|
| KL | 4.312 | 2.896    |

## 7. Related Work

Previous work attempted to exploit this same untapped resource by minimizing the Output Distribution Matching cost (Sutskever et al., 2015), which in VAE terminology, is the gap between the log marginal probability of the data and the evidence lower bound (ELBO): $KL(p(\mathbf{y})|q_\phi(\mathbf{y}|\mathbf{x}))$. A large portion of VAE research is devoted to shrinking this gap to tighten the lower bound, and so any gains made there should transfer to our framework.

(Chen et al., 2016) equipped the GAN minimax objective with a lower bound on the mutual information between the observed data ($\mathbf{x}$) and the latent code ($\mathbf{y}$ here) given $\mathbf{z}$:

$$\mathbf{I}(\mathbf{y}; \mathbf{x}' \sim p(\mathbf{x}|\mathbf{y}, \mathbf{z})) \geq \mathbb{E}_{p(\mathbf{y})p_\theta(\mathbf{x}|\mathbf{y},\mathbf{z})}\Big[\log q_\phi(\mathbf{y}|\mathbf{x})\Big] + H(\mathbf{y}) \quad (19)$$

If we fix $\mathbf{z}$ in our reverse model, this is equivalent to first term in equation 9 (sans $H(\mathbf{y})$). This objective can be also viewed as reconstruction error on the reverse model which is was motivated our proposed VAE.

Our work can also be seen as strengthening the influence of the prior distribution. By directly feeding $\mathbf{y}_u \sim p(\mathbf{y})$ to our model, we are effectively treating $p(\mathbf{y})$ as "truth", therefore, this work follows in line with that of (Makhzani et al., 2015) and (Higgins et al., 2017) where a GAN and scaled-KL-divergence term are used, respectively, to more harshly penalize deviations of the posterior from the prior.

## 8. Conclusion & Future Work

In this work, we identified a potentially untapped resource, *unfeatured* labels. We then proposed an extension to the semi-supervised variational autoencoder capable of leveraging this newfound training signal. Empirical results on two real-world problems (Mars hyperspectral imaging & LIBS spectroscopy) and a twist on a familiar generative modeling domain support the value of *unfeatured* labels and the generality of our approach.

In future work, we will investigate our model's ability to improve performance on computer vision and NLP tasks. For example, image captioning requires assigning a short (limited length) description to an image. In this scenario, images are plentiful, yet captioned images are rare. Furthermore, captions follow a very specific structure enforced by the language grammar—this allows them to be synthesized which would provide a large *untapped* label source.

# A. Appendix

## A.1. Alternative ELBOs

As mentioned in the paper, we actually treat $\mathbf{z}$ as observed in the reverse model giving us $(\mathbf{y}_u, \mathbf{z}_u)$ pairs. This results in a minor modification to the evidence lower bound in equation (10).

$$\log q_\phi(\mathbf{y}, \mathbf{z}) \geq \mathbb{E}_{p_\theta(\mathbf{x}|\mathbf{y},\mathbf{z})}\Big[\log q_\phi(\mathbf{y}|\mathbf{x}) + \log q_\phi(\mathbf{z}|\mathbf{x}, \mathbf{y}) \\ - \log p_\theta(\mathbf{x}|\mathbf{y}, \mathbf{z}) + \log q(\mathbf{x})\Big] \quad (20)$$

Another alternative is to introduce an additional network, $q_\phi(\mathbf{z}|\mathbf{y})$ into the inference process.

$$\log q_\phi(\mathbf{y}) \geq \mathbb{E}_{q_\phi(\mathbf{z}|\mathbf{y})p_\theta(\mathbf{x}|\mathbf{y},\mathbf{z})}\Big[\log q_\phi(\mathbf{y}|\mathbf{x}) + \log q_\phi(\mathbf{z}|\mathbf{x}, \mathbf{y}) \\ - \log q_\phi(\mathbf{z}|\mathbf{y}) - \log p_\theta(\mathbf{x}|\mathbf{y}, \mathbf{z}) \\ + \log q(\mathbf{x})\Big] \quad (21)$$

and encourage internal consistency within the model by minimizing

$$KL(q_\phi(\mathbf{z}|\mathbf{y}) \parallel \int_\mathbf{x} q_\phi(\mathbf{z}|\mathbf{x}, \mathbf{y})) \quad (22)$$

using Monte Carlo samples to approximate the integral.

## A.2. Network Architectures & Training Setup

### A.2.1. COMMON

- Optimizer: Adam with gradient clipping $(-1, 1)$, $\beta_1 = 0.9$, $\beta_2 = 0.999$, $\epsilon = 1e - 4$
- Monte Carlos samples to estimate expectation: 1
- Priors are all uniform distributions
- $\mathbf{x}$-discriminative loss: L2
- $\mathbf{y}$-discriminative loss: KL-divergence

### A.2.2. CRISM

- $\mathbf{z}$-dimensionality: 1
- Nonlinearities: $tanh$
- $\mathbf{x} \to \mathbf{y}$ hidden units: [5]
- $\mathbf{x} \to \mathbf{y}$ sampling distribution: diagonal Gaussian
- $\mathbf{x} \to \mathbf{y}$ output nonlinearity: softmax
- $(\mathbf{x}, \mathbf{y}) \to \mathbf{z}$ hidden units: [5]
- $\mathbf{x} \to \mathbf{y}$ sampling distribution: diagonal Gaussian
- $(\mathbf{x}, \mathbf{y}) \to \mathbf{z}$ output nonlinearity: sigmoid
- $(\mathbf{y}, \mathbf{z}) \to \tilde{\mathbf{x}}$ hidden units: [20]
- $\mathbf{x} \to \mathbf{y}$ sampling distribution: diagonal Gaussian
- $(\mathbf{y}, \mathbf{z}) \to \tilde{\mathbf{x}}$ output nonlinearity: None
- $(\alpha_f, \alpha_f^d, \alpha_r, \alpha_r^d) = (1, 1, 0.01 \text{ (0 for M2)}, 1)$
- Batch size: 100
- # of training epochs: 5000
- Learning rate: 0.003

### A.2.3. LIBS

- $\mathbf{z}$-dimensionality: 2
- Nonlinearities: $tanh$
- $\mathbf{x} \to \mathbf{y}$ hidden units: [25, 10]
- $\mathbf{x} \to \mathbf{y}$ sampling distribution: diagonal Gaussian
- $\mathbf{x} \to \mathbf{y}$ output nonlinearity: softmax
- $(\mathbf{x}, \mathbf{y}) \to \mathbf{z}$ hidden units: [5, 5]
- $\mathbf{x} \to \mathbf{y}$ sampling distribution: diagonal Gaussian
- $(\mathbf{x}, \mathbf{y}) \to \mathbf{z}$ output nonlinearity: sigmoid
- $(\mathbf{y}, \mathbf{z}) \to \tilde{\mathbf{x}}$ hidden units: [50]
- $\mathbf{x} \to \mathbf{y}$ sampling distribution: diagonal Gaussian
- $(\mathbf{y}, \mathbf{z}) \to \tilde{\mathbf{x}}$ output nonlinearity: softplus
- $(\alpha_f, \alpha_f^d, \alpha_r, \alpha_r^d) = (0.01, 10, 0.0001 \text{ (0 for M2)}, 0.0001)$
- Batch size: 100
- # of training epochs: 2000
- Learning rate: 0.01

### A.2.4. MNIST

- $\mathbf{z}$-dimensionality: 2
- Nonlinearities: softplus
- $\mathbf{x} \to \mathbf{y}$ hidden units: [500]
- $\mathbf{x} \to \mathbf{y}$ sampling distribution: Concrete distribution / Gumbel-softmax
- $\mathbf{x} \to \mathbf{y}$ output nonlinearity: None
- $(\mathbf{x}, \mathbf{y}) \to \mathbf{z}$ hidden units: [500]

- $\mathbf{x} \to \mathbf{y}$ sampling distribution: diagonal Gaussian
- $(\mathbf{x}, \mathbf{y}) \to \mathbf{z}$ output nonlinearity: sigmoid
- $(\mathbf{y}, \mathbf{z}) \to \tilde{\mathbf{x}}$ hidden units: [500]
- $\mathbf{x} \to \mathbf{y}$ sampling distribution: Bernoulli distribution
- $(\mathbf{y}, \mathbf{z}) \to \tilde{\mathbf{x}}$ output nonlinearity: None
- $(\alpha_f, \alpha_f^d, \alpha_r, \alpha_r^d) = (0.1, 1, 0.1 \text{ (0 for M2)}, 0.1)$
- Batch size: 10000
- # of training epochs: 1000
- Learning rate: 0.001